%% file: aaai25.tex
\documentclass[letterpaper]{article} % DO NOT CHANGE THIS
\usepackage{aaai25}
\usepackage{times}  % DO NOT CHANGE THIS
\usepackage{helvet}  % DO NOT CHANGE THIS
\usepackage{courier}  % DO NOT CHANGE THIS
\usepackage[hyphens]{url}  % DO NOT CHANGE THIS
\usepackage{graphicx} % DO NOT CHANGE THIS
\usepackage{natbib}  % DO NOT CHANGE THIS AND DO NOT ADD ANY OPTIONS TO IT
\usepackage{caption} % DO NOT CHANGE THIS AND DO NOT ADD ANY OPTIONS TO IT
\usepackage{algorithm}
\usepackage{algorithmic}
\usepackage{comment}

\usepackage{newfloat}
\usepackage{listings}
\DeclareCaptionStyle{ruled}{labelfont=normalfont,labelsep=colon,strut=off} % DO NOT CHANGE THIS
\floatstyle{ruled}
\newfloat{listing}{tb}{lst}{}
\floatname{listing}{Listing}
\title{Comparing LLM Text Annotation Skills: \\A Study on Human Rights Violations in Social Media Data}
\author {
    Poli Apollinaire Nemkova,\textsuperscript{\rm 1}
    Solomon Ubani,\textsuperscript{\rm 2}
    Mark V. Albert\textsuperscript{\rm 1}
}
\affiliations {
    \textsuperscript{\rm 1}College of Computer Science and Engineering, University of North Texas, USA\\
    \textsuperscript{\rm 2}Intuit, USA\\
    poli.nemkova@unt.edu, solomon\_ubani@intuit.com, mark.albert@unt.edu
}

\usepackage[russian, english]{babel}
\usepackage{graphicx}

\begin{document}

\maketitle

\input{Abstract}

\section{Introduction}
\input{Introduction}

\section {Literature Review}
\input{LitReview}

\section{Experiment Design}
\input{Experiment_Design}

\subsection{Dataset}
\input{Dataset}

\subsection{Method}
\input{method}

\section{Results}

\input{results}

\section{Error Analysis and Discussion}
\input{Error_Analysis}

%\section{Discussion}
%\input{AnonymousSubmission/LaTeX/Discussion}

\section{Limitations}
\input{Limitation}

\section{Conclusion}
\input{Conclusion}

\bibliography{aaai25}

\end{document}

%% file: Abstract.tex
\iffalse
\begin{abstract}

In the era of increasingly sophisticated natural language processing (NLP) systems, large language models (LLMs) have demonstrated remarkable potential for diverse applications, including tasks requiring nuanced textual understanding and contextual reasoning. This study investigates the capabilities of multiple state-of-the-art LLMs—GPT-3.5, GPT-4, LLAMA3, Mistral 7B, and Claude-2—for zero-shot and few-shot annotation of a complex textual dataset comprising social media posts in Russian and Ukrainian. Specifically, the focus is on the binary classification task of identifying references to human rights violations within the dataset.

To evaluate the effectiveness of these models, their annotations are compared against a gold standard set of human double-annotated labels across 1000 samples. The analysis includes assessing annotation performance under different prompting conditions, with prompts provided in both English and Russian. Additionally, the study explores the unique patterns of errors and disagreements exhibited by each model, offering insights into their strengths, limitations, and cross-linguistic adaptability.

By juxtaposing LLM outputs with human annotations, this research contributes to understanding the reliability and applicability of LLMs for sensitive, domain-specific tasks in multilingual contexts. It also sheds light on how language models handle inherently subjective and context-dependent judgments, a critical consideration for their deployment in real-world scenarios.
\end{abstract}
\fi

\begin{abstract}

Large language models (LLMs) have shown promise in tasks requiring nuanced textual understanding. This study evaluates state-of-the-art LLMs—GPT-3.5, GPT-4, LLAMA3, Mistral 7B, and Claude-2—for annotating a multilingual dataset of social media posts in Russian and Ukrainian, focusing on identifying human rights violations. Model performance is compared against a human-annotated gold standard across 1000 samples, with analysis of zero-shot and few-shot prompting in English and Russian.

Key findings show GPT-4.0 excels in precision and recall, making it ideal for high-stakes tasks. Aligning prompt language with dataset language improves performance, particularly for open-source models like LLaMA3 and Mistral-7B, which benefit from few-shot prompting but require additional validation for sensitive applications. Models, like humans, tend to struggle with ambiguous cases, particularly those involving indirect or nuanced references. Closed-source models demonstrate robust zero-shot capabilities with marginal gains from few-shot setups.

This study highlights the importance of linguistic alignment, prompt strategy, and model selection, offering practical insights for deploying LLMs in multilingual, high-impact domains like human rights monitoring.

\end{abstract}

%% file: Introduction.tex
The emergence of large language models (LLMs) has revolutionized natural language processing (NLP), pushing the boundaries of what machines can achieve in understanding, generating, and classifying text. Models such as GPT-3.5, GPT-4, Claude-2, and open-source alternatives like LLaMA and Mistral-7B have demonstrated extraordinary capabilities across a wide range of applications, from translation to content summarization. However, the deployment of LLMs for high-stakes, domain-specific tasks, such as identifying references to human rights violations in multilingual social media data, remains a critical yet underexplored area of research.

The domain of human rights monitoring presents unique challenges. It involves the analysis of contextually rich, multilingual, and emotionally charged texts, often requiring nuanced comprehension and interpretation of implicit or indirect references. Traditionally, these tasks have been conducted by human annotators, whose expertise ensures high accuracy but incurs significant costs in time and resources. LLMs offer an unprecedented opportunity to automate and scale such tasks, but their ability to handle the inherent subjectivity and complexity of human rights-related data is not well understood. Moreover, questions remain about how factors such as prompt design, linguistic alignment, and model architecture influence performance.

In this study, we systematically evaluate the performance of leading LLMs and open-source alternatives on the binary classification task of detecting references to human rights violations in social media posts written in Russian and Ukrainian. Using a dataset curated from public Telegram channels, we compare models under two prompting configurations—zero-shot and few-shot learning—and investigate the effect of prompt language (English vs. Russian) on model performance. Additionally, we examine differences between closed-source models, such as GPT-4 and Claude-2, and open-source models, such as LLaMA and Mistral-7B, to understand the trade-offs between cost, accessibility, and performance.

\textbf{Key contributions of this study: } 
\begin{itemize}
    \item \textbf{Comprehensive Benchmarking Across Models and Prompts:} We provide a comparative analysis of closed- and open-source LLMs under different prompting styles (zero-shot vs. few-shot) and languages (English vs. Russian), highlighting performance gaps and strengths.

    \item \textbf{Insights Into Prompting Strategies:} The study demonstrates that open-source models benefit significantly from few-shot prompting with examples, whereas closed-source models exhibit robust performance across both zero-shot and few-shot settings.

    \item \textbf{Impact of Linguistic Alignment:} We show that aligning the prompt language with the data language (e.g., Russian prompts for Russian texts) consistently improves model performance across architectures, emphasizing the importance of multilingual capability in real-world tasks.

    \item \textbf{Error Analysis and Practical Guidance:} Through a detailed analysis of error types and model disagreements with human annotations, we provide actionable insights for selecting and optimizing LLMs in high-context, sensitive domains.

    \item \textbf{Cost-Effectiveness of Open-Source Models:} Despite their lower baseline performance, open-source models demonstrate strong potential when optimized with appropriate prompting strategies, making them viable alternatives for resource-constrained scenarios.

\end{itemize}
This work advances the understanding of LLM capabilities in multilingual, high-stakes applications and offers practical guidelines for leveraging these models effectively in real-world human rights monitoring tasks. By addressing critical gaps in model evaluation and exploring the interplay of architecture, prompt design, and language alignment, this study lays the groundwork for future research and development in this domain.

%% file: LitReview.tex
The emergence of Large Language Models (LLMs) has transformed Natural Language Processing (NLP), particularly in text annotation, enabling automation at a scale and speed unattainable through traditional methods. This review explores the comparative effectiveness of LLMs and human annotators in text annotation tasks\cite{10460013, nasution2024chatgpt}, highlighting their strengths, limitations, and the pressing need for further research into their labeling performance, especially for complex and high-context tasks.

\textbf{Efficiency and Cost-Effectiveness.}
LLMs have demonstrated substantial potential to reduce the time and cost associated with text annotation. For instance, integrating LLMs with human expertise in medical information extraction has been shown to significantly reduce manual effort while maintaining high accuracy, facilitating the rapid creation of labeled datasets \cite{goel2023llms}. Similarly, the CoAnnotating framework leverages LLMs to complement human annotators by allocating annotation tasks based on uncertainty, achieving a 21\% performance improvement over random baselines \cite{li2023coannotating}.

However, these gains are task-dependent. For more nuanced annotations, such as legal or ethical contexts, the costs of fine-tuning or prompt engineering can be substantial, offsetting the initial efficiency benefits \cite{brown2020language, bommasani2021opportunities}. This highlights a critical need to understand where LLMs can outperform or meaningfully complement traditional annotation workflows.

\textbf{Accuracy and Reliability. }
While LLMs excel in generalization, particularly in zero-shot and few-shot learning, their performance varies significantly across tasks and datasets. For example, ChatGPT-4 has outperformed both expert classifiers and crowd workers in annotating political Twitter messages, demonstrating higher accuracy and consistency \cite{tornberg2023chatgpt}. Similarly, GPT-3 and GPT-4 have shown strong performance in summarizing legal documents and medical records, often rivaling domain experts \cite{Šavelka2023Unlocking, Shaib2023Summarizing, Takagi2023Performance}.

However, studies have also highlighted limitations. For instance, LLMs frequently underperform on tasks requiring deep contextual understanding or domain-specific expertise, often lagging behind fine-tuned smaller models trained on expert-annotated data \cite{plaza2023leveraging, pangakis2023automated}. Furthermore, while LLMs achieve high accuracy in standard tasks, their susceptibility to generating false positives or negatives in edge cases or nuanced scenarios remains a critical limitation.

\textbf{Domain-Specific Knowledge. }
Human annotators possess domain-specific knowledge that is often critical for accurate text annotation. In tasks requiring subtle contextual judgments, such as medical diagnosis or human rights monitoring, small models trained on curated datasets have been shown to outperform general-purpose LLMs \cite{lu2023human}. These findings underscore the limitations of pre-trained LLMs when applied to tasks requiring specialized knowledge and suggest that domain adaptation remains a key challenge.

Research also indicates that while in-context learning (e.g., few-shot prompting) improves LLM performance in specialized domains, it is not always sufficient to match the nuanced understanding of human annotators \cite{shin-etal-2020-autoprompt}. This highlights a need for further exploration into how prompting strategies can bridge this gap, particularly in complex labeling tasks.

\textbf{Bias and Ethical Considerations}
The use of LLMs in text annotation raises significant concerns about bias and ethical implications. While models like ChatGPT-4 have shown reduced bias compared to human annotators in some tasks \cite{tornberg2023chatgpt}, they are far from neutral. Biases inherent in training data often manifest in annotations, particularly in sensitive or controversial domains such as politics, law, or human rights \cite{fisher2024biased, bender2021dangers, Zimmer2010“But}.

Auditing frameworks such as ALLURE aim to address these challenges by incorporating failure cases into the evaluator through in-context learning, highlighting the importance of iterative improvement \cite{hasanbeig2023allure}. However, such frameworks remain experimental, and the broader challenges of transparency and accountability in LLM-based annotation persist \cite{bommasani2021opportunities}.

\textbf{Open-Source vs. Proprietary Models. }
Open-source models such as LLaMA,   Mistral,  or FLAN-T5 have emerged as viable alternatives to proprietary systems like GPT-4, offering cost-effectiveness, transparency, and superior data protection \cite{touvron2023llama, raffel2020exploring, jiang2023mistral}. Studies have shown that open-source LLMs can outperform crowd-sourced services like Amazon MTurk in specific tasks and even achieve competitive performance against proprietary models in domains like sentiment analysis and legal classification \cite{wong2024comparative}.

However, open-source models are often more sensitive to prompting strategies and lack the robust pretraining data that benefits proprietary systems, particularly in multilingual and context-sensitive tasks. This creates a trade-off between cost and performance that must be better understood through systematic benchmarking \cite{bommasani2021opportunities}.

\textbf{Need for Research on LLM Labeling. }
Despite their promise, the use of LLMs in labeling tasks remains underexplored, particularly for complex scenarios requiring nuanced understanding. Most existing studies focus on standard tasks (e.g., sentiment analysis, named entity recognition), where LLMs often achieve near-human performance. However, tasks involving ambiguity, indirect references, or ethical considerations, such as detecting human rights violations in multilingual social media posts, are far more challenging and less studied \cite{artstein2008inter, Törnberg2023ChatGPT-4}.

Understanding how LLMs navigate these challenges—particularly in zero-shot and few-shot settings—is critical for assessing their suitability for real-world applications. This study addresses this gap by evaluating LLM performance in labeling complex, multilingual datasets, exploring the interplay of prompt design, linguistic alignment, and model architecture. By providing insights into model strengths, limitations, and error patterns, this work contributes to the growing body of research on optimizing LLMs for domain-specific annotation tasks.

%% file: Experiment_Design.tex
\textbf{General Framework}

 This study assesses the capability of large language models (LLMs) to annotate a dataset of Russian and Ukrainian social media posts for references to human rights violations, comparing their performance against a gold-standard dataset created by human annotators. The dataset underwent a rigorous double-annotation process, with disagreements resolved by a senior adjudicator, resulting in a finalized benchmark label set.

The initial Cohen's Kappa score (before adjudication) for annotator agreement was 0.63, indicating substantial agreement. Out of 1000 samples, human annotators disagreed on 184 instances, which were subsequently resolved by the adjudicator. Despite the substantial agreement, the annotation task remains challenging for humans.

By comparing the outputs of LLMs with the human-labeled data, this study explores the extent to which these models can replicate human annotations in a multilingual, context-sensitive domain. Furthermore, the study investigates the performance of LLMs on the 816 fully agreed-upon samples and the 184 samples with initial disagreements separately, providing a nuanced evaluation of their capabilities.

%% file: Dataset.tex
To evaluate model annotation performance, we used a sample from a larger Human Rights Violations (HRV) dataset \cite{Nemkova2023DetectingHR}. The dataset comprises social media posts collected from public Telegram\footnote{https://telegram.org/} news channels, primarily focused on the Russia-Ukraine conflict. These posts are written predominantly in Russian, with a smaller proportion in Ukrainian (966 post in Russian, 34 in Ukrainian). Telegram has been increasingly used for public discourse and news dissemination, particularly in the context of conflict reporting \footnote{https://time.com/6158437/telegram-russia-ukraine-information-war/}.

Each post was annotated for the presence or absence of references to situations falling under the category of human rights violations. The annotation process involved two trained volunteers who are native speakers of either Russian or Ukrainian, with proficiency in understanding both languages. The label is binary: the positive class indicates the presence of a reference to a human rights violation (HRV), while the negative class covers posts where HRV references are absent or unclear. In cases where the two annotators disagreed, a third annotator with domain-specific expertise adjudicated the final label. Such adjudication processes are widely used in NLP to resolve disagreements and ensure high-quality ground truth annotations \cite{artstein2008inter}.

Inter-annotator agreement was measured using Cohen’s Kappa \cite{cohen1960coefficient} with value 0.63, which indicated substantial  agreement between the two primary annotators. This level of agreement reflects the inherent complexity of the task, as detecting references to human rights violations often involves interpreting nuanced or implicit contextual cues. However, the adjudication process ensured the creation of a high-quality gold-standard dataset suitable for benchmarking.

For this study, a sample of 1000 posts was selected from the dataset. The sample is moderately imbalanced, with 517 of posts labeled as belonging to the positive class (HRV present) and the remaining 483 labeled as negative (no HRV or unclear). Class imbalance is a common challenge in such tasks and can impact model performance \cite{japkowicz2002class}. This curated sample served as the benchmark against which the LLMs' annotation performance was evaluated.

%Examples of posts from the dataset and their corresponding %human annotations are provided in Table~X. These examples %illustrate the types of content analyzed and the binary %labels assigned by the human annotators, providing %transparency into the annotation process.

%% file: method.tex
We evaluated the following LLMs to capture a diverse range of capabilities, including both state-of-the-art proprietary systems and open-source models:

\textbf{GPT-4.0 and GPT-3.5 Turbo (OpenAI)}:
GPT-4.0 represents the current cutting-edge in language modeling, demonstrating superior performance across a range of multilingual NLP tasks \cite{openai2023gpt4}. GPT-3.5 Turbo, while less powerful, is more cost-effective and serves as a strong baseline for comparison in terms of practical deployment.

\textbf{Claude-2 (Anthropic)}:
Claude-2 is designed with a focus on safety and alignment, particularly relevant in domains involving sensitive or ethical considerations \cite{anthropic2023claude}.

\textbf{LLaMA-3.2-1B (Meta)}:
This smaller-scale open-source model \cite{touvron2023llama} was included to evaluate the feasibility of using lightweight, publicly available models for multilingual annotation tasks.

\textbf{Mistral-7B}:
Another open-source model, Mistral-7B \cite{mistral2023release}, was selected for its promising performance on various NLP tasks while maintaining a smaller computational footprint.

The selection of models reflects a balance between high-performing proprietary systems and resource-efficient open-source models, providing a comprehensive perspective on LLM performance across different architectures and scales.

\textit{Experimental Setup. }
All experiments were conducted in Python in a Google Colab environment to ensure accessibility and reproducibility. High RAM A100 was utilized. None of the models were fine-tuned on the dataset. Instead, we utilized their zero-shot and few-shot capabilities to assess their out-of-the-box performance, a common real-world scenario where task-specific fine-tuning is infeasible due to data limitations or computational constraints.

\textit{Prompting and Evaluation. } \footnote{All prompts used in this experiment are available at the experiment GitHub link: https://GitHub.com/PoliNemkova/LLM\_labeling\_skills} 
Two prompting strategies were employed:
\begin{itemize}
    \item \textit{Zero-shot prompting:} Direct task instructions without providing examples. This evaluates the models’ ability to generalize to the task based on pretraining alone.
    \item \textit{Few-shot prompting:} Instructions augmented with multiple labeled examples. This approach leverages in-context learning to improve task-specific alignment \cite{brown2020language}.
\end{itemize}
Prompts were tested in both English and Russian, aligning with the linguistic characteristics of the dataset. Russian prompts generally outperformed English prompts across models, demonstrating the importance of prompt language alignment in multilingual tasks. The best-performing prompts for each model and setting are detailed in the GitHub.

Performance was evaluated using precision, recall, F1 score, and accuracy, with a focus on balancing sensitivity and specificity, particularly given the dataset's moderate positive class imbalance.

\textit{Reproducibility. }
To ensure replicability, all experiments were conducted with fixed random seeds. The seeds, along with detailed prompt designs and hyperparameter settings, are provided in the project GitHub. This adherence to reproducibility standards aligns with best practices in the NLP community \cite{pineau2021improving}.

%% file: results.tex
\begin{figure*}[!t]
    \centering
    \includegraphics[width=0.7\textwidth]{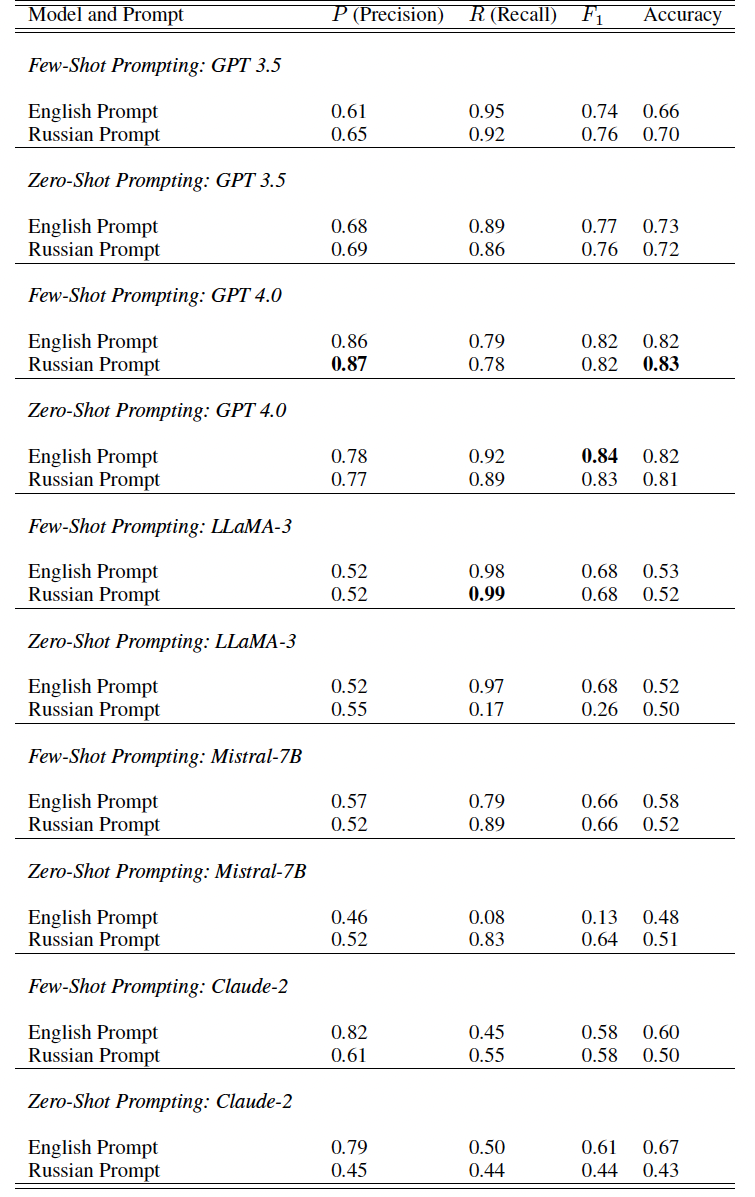}
    \caption{Performance metrics for GPT 3.5, GPT 4.0, LLaMA-3, Mistral-7B, and Claude-2 in Few-Shot and Zero-Shot Prompting scenarios using English and Russian prompts. Bold values indicate the highest performance for each column across all rows.}
    \label{fig:rezzzult}
\end{figure*}

%%%
The performance metrics for all models are detailed in Figure \ref{fig:rezzzult} . Among the evaluated models, \textbf{GPT-4.0} demonstrated the highest overall performance, particularly excelling in zero-shot prompting with English prompts, where it achieved the top F1 score of 0.84 and an accuracy of 0.82. This performance highlighted its strong balance between precision (0.78) and recall (0.92). Similarly, \textbf{GPT-3.5} performed well, particularly in few-shot prompting with Russian prompts, achieving an F1 score of 0.76 and an accuracy of 0.70, showcasing its capability in scenarios where high recall (0.92) is essential.

In contrast, \textbf{LLaMA-3} and \textbf{Claude-2} exhibited limitations. LLaMA-3 struggled with consistently low accuracy and F1 scores, particularly in zero-shot Russian prompting (F1 = 0.26, Accuracy = 0.50). Claude-2 showed imbalanced performance, achieving moderate precision but relatively lower recall across all setups. For example, in few-shot prompting with English, it reached an F1 score of 0.58 but had recall as low as 0.45, signaling challenges in achieving generalizable results.

The results emphasize the importance of aligning prompt language with the dataset. Russian prompts generally outperformed English prompts across models, particularly with GPT-3.5 and GPT-4. Few-shot prompting tended to enhance recall, as seen in GPT-3.5 and Mistral-7B, but sometimes led to trade-offs in precision. Meanwhile, zero-shot prompting with GPT-4 delivered the most balanced and robust outcomes, making it a reliable choice for diverse task requirements.

%% file: Error_Analysis.tex
\subsection{Model Performance Overview} 
\iffalse
We evaluated a diverse set of language models, including both paid (GPT-3.5, GPT-4.0, Claude2) and open-source models (Llama3, Mistral7B), for their ability to identify references to human rights violations in a multilingual dataset. The dataset was annotated with binary labels, and the models were tested under different configurations of prompt language (English vs. Russian) and prompting style (zero-shot vs. few-shot).
\fi

\input{Discussion}

\textbf{Model Disagreement and Error Analysis}
This section examines where and why the models diverged in their predictions, focusing on key sources of disagreement and common error patterns.
\\
\textit{Error analysis identified key challenges for the models:}
\begin{itemize}
    \item \textit{False Positives:} Frequently occurred in cases involving events such as reports of military activity or protests, where human rights violations were implied but not explicitly mentioned.
    \item \textit{False Negatives:} Often observed in posts containing indirect references or systemic violations, requiring deeper contextual understanding.
    \item \textit{Disagreement on Ambiguity}: Posts with the highest disagreement involved ambiguous language or context, where models diverged in their interpretation of human rights implications.
\end{itemize}

\input{ablation}
\\

\textbf{Key Findings}
\begin{enumerate}
    \item \textbf{Advanced Models Perform Better}: GPT-4.0 should be prioritized for tasks requiring high precision and recall. Its robust performance across multiple settings demonstrates the value of state-of-the-art LLMs.
 
    \item \textbf{Alignment of Prompt Language Improves Performance}: Models consistently performed better with prompts in Russian, reflecting the importance of linguistic alignment when working with multilingual datasets.
   
    \item \textbf{Few-Shot Prompting Boosts Open-Source Model Performance}: Open-source models showed substantial improvements with few-shot prompting compared to zero-shot prompting. In contrast, closed-source models like GPT-4.0 and GPT-3.5 displayed strong zero-shot capabilities, with marginal gains in few-shot settings.
  
    \item \textbf{Additional Validation for Open-Source Models}: The moderate precision rates of open-source models, such as Mistral-7B Few-Shot (Russian) and LLaMA-3 Few-Shot (Russian), warrant an additional manual review step to mitigate false positives in sensitive applications.
\end{enumerate}

\textbf{Future Directions}
Future work should focus on enhancing the multilingual capabilities of smaller LLMs like LLaMA and Mistral through \textit{fine-tuning} strategies or domain-specific pretraining. Additionally, investigating advanced \textit{prompt engineering} techniques and dataset augmentation could help mitigate the performance gaps of underperforming models, especially in context-sensitive and low-resource scenarios. Expanding experiments to include a \textit{broader range of languages} will also provide valuable insights into the adaptability and generalization of these models across diverse linguistic contexts.

%% file: Discussion.tex
This study evaluated the performance of multiple language models (LLMs) on the task of detecting human rights violations in a multilingual dataset comprising Russian and Ukrainian social media posts. The dataset’s positive class ratio reflects a moderate imbalance, necessitating careful consideration of both recall (to minimize false negatives) and precision (to limit false positives). The analysis underscores the critical impact of model architecture, prompting strategy, and prompt language alignment on model performance.

\textbf{Best Performing Models}

\textbf{GPT-4.0} emerged as the strongest performer across all settings, demonstrating consistent superiority in both zero-shot and few-shot scenarios. In particular, zero-shot prompting with English prompts achieved the highest overall F1 score (0.84) and accuracy (0.82), showcasing GPT-4’s exceptional ability to generalize effectively from minimal context. This performance reflects a robust balance between precision (0.78) and recall (0.92), which is critical for tasks involving human rights violations where both false positives and false negatives carry significant consequences.

\textbf{GPT-3.5} also showed competitive performance, particularly in few-shot prompting with Russian prompts (F1 = 0.76, Recall = 0.92), demonstrating its ability to capture a high proportion of true positives. While GPT-3.5 fell short of GPT-4’s overall performance, its cost-efficiency and recall-centric behavior in certain settings make it a strong candidate for tasks prioritizing coverage.

Both GPT-4 and GPT-3.5 exhibited a pronounced benefit from using Russian prompts, suggesting their enhanced alignment with the linguistic and contextual features of the dataset. These results emphasize the importance of prompt language matching, particularly when working with models trained primarily on multilingual or English-centric corpora.

\textbf{Underperforming Models}

\textbf{LLaMA-3} and \textbf{Claude-2} consistently underperformed across most settings. LLaMA-3 struggled with balancing precision and recall, achieving an F1 score as low as 0.26 and an accuracy of 0.50 in zero-shot Russian prompting. While its recall improved dramatically in few-shot settings with Russian prompts (0.99), the corresponding drop in precision (0.52) undermines its practical applicability. Similarly, Claude-2 exhibited imbalanced behavior, achieving moderate F1 scores in few-shot English prompting (F1 = 0.58) but struggling with recall (0.45), indicating an inability to generalize effectively.

These results highlight that both models lack the robustness required for complex tasks in multilingual and context-sensitive domains. Notably, Claude-2 and LLaMA-3 were particularly sensitive to prompt design, and their performance varied significantly based on the prompting language, with both models generally performing worse on Russian prompts.

\textbf{Prompting Strategy}

The results reveal notable differences in model performance based on prompting strategy. \textbf{Few-shot prompting} often achieved higher recall across models, as seen with GPT-3.5 (Russian Few-Shot Recall = 0.92) and LLaMA-3 (Russian Few-Shot Recall = 0.99). This indicates that few-shot prompting can provide models with a stronger inductive bias toward identifying true positives in imbalanced datasets. However, the trade-off was reduced precision in many cases, as seen with LLaMA-3 (Russian Few-Shot Precision = 0.52).

In contrast, \textbf{zero-shot prompting} tended to produce more balanced results, yielding higher F1 scores and accuracy. GPT-4, for instance, achieved its best performance in the zero-shot English prompt setting (F1 = 0.84), reflecting its ability to effectively leverage minimal contextual information without overfitting to specific patterns in the prompt.

\textbf{Implications for Multilingual NLP}

The significant performance gap between Russian and English prompts across all models reinforces the \textit{critical role of linguistic alignment in multilingual tasks}. Even models designed with multilingual capabilities, such as GPT-4, displayed notable improvements when prompted in Russian, which aligns directly with the dataset’s language. This suggests that for multilingual or non-English tasks, careful prompt design in the target language is essential to achieving optimal results.

Furthermore, the stark underperformance of Claude-2 and LLaMA-3 highlights the limitations of smaller or less advanced models in generalizing effectively for non-English, context-sensitive tasks. This finding raises important questions about the scalability and robustness of smaller LLMs in real-world applications, particularly when deployed in low-resource or non-English domains.

%% file: ablation.tex
\textbf{Ablation Study}\\
Two of the best-performing models, GPT-4.0 (proprietary) and LLaMA-3 (open-source), were evaluated on two distinct subsets of the original dataset: one where human annotators were in full agreement on the labels, and another where annotators disagreed, requiring adjudication. The goal of this analysis was to determine whether large language models (LLMs) face similar challenges to those encountered by humans. Results are shown in Figure 2. \\

\begin{figure*}[!t]
    \centering
    \includegraphics[width=1\textwidth]{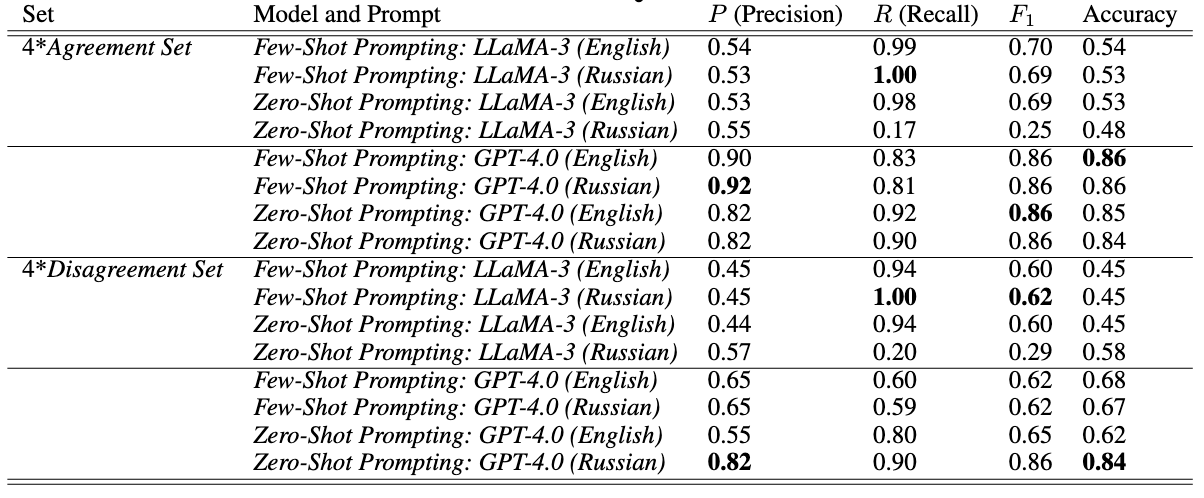}
    \caption{Performance metrics for LLaMA-3 and GPT-4.0 on the agreement and disagreement sets in Few-Shot and Zero-Shot Prompting scenarios using English and Russian prompts. Bold values indicate the highest performance for each column across all rows.}
    \label{fig: ablation}
\end{figure*}

\textit{Do Models Face the Same Challenges as Humans?}

\textbf{1- Shared Struggles}
\begin{itemize}
    \item Disagreement cases likely involve ambiguous language, edge cases, or subtle contextual nuances that make labeling more challenging.
    \item Both LLaMA-3 and GPT-4.0 exhibit significant drops in Precision on these cases, suggesting an increased likelihood of false positives when ambiguity is present.
\end{itemize}

\textbf{2 - GPT-4.0 Shows Greater Resilience}
- GPT-4.0 demonstrates a smaller decrease in F1 Score and Accuracy compared to LLaMA-3, indicating that it handles disagreement cases more effectively.
- This suggests that GPT-4.0 may better capture nuanced patterns and context, reducing errors that LLaMA-3 struggles with in these challenging scenarios.

%% file: Limitation.tex
Despite promising findings, this study has several limitations that should be addressed in future work.

\textbf{Language Coverage}:
The dataset focuses on Russian and Ukrainian posts, which limits generalizability to other languages or dialects. Although Russian prompts generally outperformed English prompts, additional analysis is needed to understand model performance on datasets with multiple or less-resourced languages.

\textbf{Prompt Design Limitations:}
The study used standardized zero-shot and few-shot prompts, but the potential of prompt engineering remains underexplored. Optimizing prompt structure, content, or complexity could further enhance performance, particularly for underperforming models.

\textbf{Sample Size:}
The study utilized a random sample of 1000 posts to evaluate model performance. While a larger dataset would enable more robust and generalizable conclusions, the computational cost associated with employing models such as GPTs or Claude makes the use of an expanded dataset significantly more expensive.

\textbf{Computational Costs:}
While GPT-4 and GPT-3.5 achieved strong results, their computational demands and associated costs may limit their scalability in real-world applications. Smaller models like LLaMA-3.2-1B, despite their lower performance, offer cost advantages that could be valuable in resource-constrained environments.

%% file: Conclusion.tex
This study systematically evaluates the performance of both paid and open-source language models for identifying references to human rights violations in a multilingual dataset. By analyzing the effects of prompting style, language alignment, and model architecture, we provide actionable insights for selecting and optimizing models under varying resource constraints. 

Our findings reveal that paid models, particularly GPT-4.0, excel in both precision and recall, demonstrating robust generalization across languages and task configurations. Open-source models, such as Llama3 and Mistral7B, perform competitively with carefully designed prompts, especially in few-shot and language-aligned settings. However, their sensitivity to context and higher error rates underscore the need for task-specific fine-tuning.

The disparity in performance between Russian and English prompts highlights the importance of language alignment in multilingual tasks. Furthermore, error analysis shows that ambiguity and indirect references pose significant challenges, leading to model disagreements and underscoring the complexity of the human rights domain.

These insights offer practical guidance for practitioners: leveraging few-shot prompting for open-source models in low-budget settings and prioritizing GPT-4.0 for high-stakes applications where accuracy is paramount. Our study paves the way for future work to refine LLMs for nuanced, high-impact tasks such as monitoring and analyzing human rights violations globally.